# Towards AGI: A Pragmatic Approach Towards Self Evolving Agent

Indrajit Kar
ORCID: 0000-0002-9539-6719

Sammy Zonunpuia
sammy@rastrai.com

Zonunfeli Ralte
ORCID: 0009-0000-3282-3368

*Abstract*: Large Language Model (LLM) based agents are powerful yet fundamentally static after deployment, lacking the ability to autonomously expand capabilities, generate new tools, or evolve their reasoning. This work introduces a hierarchical self-evolving multi-agent framework that integrates a Base LLM, an operational SLM agent, a Code-Generation LLM, and a Teacher-LLM to enable continuous adaptation. The workflow begins with the agent attempting a task using reasoning and existing tools; if unsuccessful, it escalates to tool synthesis through the Code-Gen LLM, and when failures persist, it triggers an evolution phase using Curriculum Learning (CL), Reward-Based Learning (RL), or Genetic Algorithm (GA) evolution. Using the TaskCraft dataset rich in hierarchical tasks, tool-use traces, and difficulty scaling we evaluate these paradigms. CL delivers fast recovery and strong generalization, RL excels on high-difficulty tasks, and GA offers high behavioral diversity. Across all settings, evolved agents outperform their originals, demonstrating robust, autonomous, self-improving agentic evolution.

*Self-evolving agents; Large Language Models (LLMs); Curriculum Learning (CL); Reward-Based Learning (RL); Genetic Algorithm (GA) evolution; Multi-agent systems; Tool-augmented reasoning; Code-generation LLMs; Autonomous adaptation; TaskCraft dataset; Agentic workflows; Self-improving AI; Capability evolution; Hierarchical orchestration.*

## I. INTRODUCTION

Large Language Models (LLMs) have undergone a rapid transition from passive text generators to autonomous agents capable of planning, tool use, and memory-driven decision-making. Early LLM-powered agents were inherently static: their architectures, prompts, and workflows remained fixed after deployment, limiting their ability to adapt to new tasks or evolving environments. This limitation has been highlighted extensively in recent studies on agentic system design, which argue that the next paradigm shift must enable agents to continuously adapt and optimize themselves over time [14].

The LLM agent life cycle initialization, execution, monitoring, collaboration, and termination introduced structured workflows that improved coordination and reliability among agents. However, these agents still primarily operate as fixed-function components. To overcome this rigidity, recent research emphasizes the emergence of **self-evolving learning agents**, wherein the agent not only executes workflows but also refines them through feedback loops. Such agents leverage mechanisms like self-reflection, reward-based adaptation, trajectory improvement, and curriculum learning to autonomously enhance their competence without relying exclusively on static training corpora.

Recent work proposes a unified conceptual framework for self-evolving agents comprising four elements: **system inputs**, **the agent system**, **the environment**, and **optimizers** [14]. This framework underscores that effective autonomous evolution requires dynamic integration of memory, workflows, tools, and optimization mechanisms, including prompt evolution, tool augmentation, workflow adaptation, multi-agent collaboration, and domain-specific skills. It also highlights critical open challenges such as ensuring safe evolution, preventing misalignment, managing data drift, and detecting emergent behaviours when agents self-modify.

Our work advances this vision by introducing a hierarchical, multi-agent optimization system where learning, reasoning, and adaptation arise from continuous inter-agent communication and feedback-driven evolution. In our architecture, a Base LLM orchestrates problem decomposition and agent routing, a Code-Generation LLM synthesizes new tools and functional modules on demand, a Teacher LLM drives curriculum- or reward-based optimization, and an operational agent (SLM) performs task-level reasoning. Together, these components form a dynamic ecosystem: when an agent fails to solve a task, it escalates the issue; the Code-Gen LLM generates higher-order tools to close capability gaps; and the Teacher LLM drives iterative improvement through curriculum or reinforcement-guided evolution.

A distinguishing aspect of our system is its ability to replicate, upgrade, and retire agents. When challenges exceed current agent capabilities, the system creates a cloned agent dedicated to learning through curated curricula, reward-based signals, or genetic-algorithm-inspired evolution. Once the clone surpasses the performance of the original, it replaces it, achieving a form of lifelong self-evolution. This adaptive mechanism moves beyond static deployment models, enabling agents to autonomously expand their toolsets, optimize decision-making processes, and evolve their behaviors over time.

By integrating dynamic tool generation, curriculum-driven evolution, multi-agent coordination, and verifiable improvement mechanisms, our architecture operationalizes the paradigm of self-evolving agentic systems, addressing fundamental gaps identified in prior literature [1–14]. This represents a step toward scalable, safe, and continuously improving autonomous AI systems capable of long-term reasoning and adaptation.

## II. RELATED WORKS

The Research on tool-augmented reasoning, self-improving language models, and multi-agent LLM architectures provides the foundational basis for our proposed system. Early advances in tool-integrated reasoning demonstrate that the effectiveness of an autonomous agent increases significantly when natural-language reasoning is coupled with external computational resources. Gou et al. [1] established this principle through a tool-integrated agent capable of interleaving reasoning with real-time tool calls, while the iterative self-reflective paradigm introduced by Zelikman et al. [2] showed that agents can improve by generating and refining their own rationales. These works directly motivate the first phase of our architecture, where an agent attempts a task independently, selects tools based on internal reflection, and escalates unresolved tasks to a higher-order controller.

The integration of code synthesis and execution as part of agent reasoning is further supported by Ni et al. [3] and Yin et al. [4], whose methods illustrate that models equipped with execution-aware reasoning and code-nested solution generation achieve substantially higher reliability. These findings align with the role of the Code-Gen LLM in our workflow, which produces executable tools that can be hot-deployed by the agent to extend its functional capabilities.

Recent progress in LLM-driven multi-agent system generation reinforces the feasibility of dynamic agent orchestration. Ye et al. [5] demonstrated that multi-agent workflows can be generated directly from natural-language queries, providing methodological grounding for the orchestration responsibilities we assign to the base LLM. Complementary work on reward-based self-improvement, particularly the self-rewarding mechanisms of Yuan et al. [6] and the verifiable reward pipelines in Lambert et al. [7], informs the design of our evolutionary loop, where an agent escalates unsolved tasks for reward or curriculum-driven adaptation.

Curriculum learning and trajectory-based improvement frameworks also influence the structure of our evolutionary stage. Jiao et al. [8] introduced planning-centric reasoning through trajectory reconstruction and process rewards, providing a template for our teacher-LLM's curriculum generation. Similarly, the exploration–critique–optimization framework of Putta et al. [9] demonstrates how guided search and iterative preference learning enhance agent autonomy, offering techniques applicable to our retry and refinement cycles.

The utility of synthetic data for enabling large performance jumps without human supervision has been strongly established in mathematical and logical reasoning domains. Xin et al. [10] showed the impact of large-scale synthetic proof data, which parallels our generation of synthetic curriculum tasks for agent replication. Liu et al. [11] extended self-evolving training principles to multimodal contexts, providing insights into reward modeling and prompt design relevant to our generalizable training pipeline.

Finally, recent advancements in fully self-generated learning signals highlight the viability of autonomous agent evolution without external datasets. Zhao et al. [12] presented a zero-data reasoning framework where agents generate tasks, solutions, and verifiable signals entirely internally, while Huang et al. [13] proposed a challenger–solver dynamic that enables continuous self-curriculum construction. These studies directly support the final evolutionary step of our framework, where a replicated agent is trained through internally constructed tasks and verifiable reward mechanisms before replacing the original.

Extensions to curriculum design and data-efficient task selection emphasize adaptive task sampling as a core mechanism for scalable agent improvement. Chen et al. introduced a Self-Evolving Curriculum (SEC) that frames curriculum selection as a bandit problem, adaptively prioritizing intermediate-difficulty tasks to maximize learning gain during concurrent policy fine-tuning [15]. SEC's adaptive task prioritization informs our Teacher LLM's curriculum scheduler: rather than static curricula, we adopt bandit-style selection to focus clone training on tasks with the highest marginal benefit. Similarly, WebRL demonstrates how self-evolving online curricula can bootstrap web-agent competence by generating new tasks from failed interactions and using an adaptive reward model for RL updates [16]; this validates our approach to autonomously generate and validate web-style or environment-specific curriculum items when agents encounter novel external interfaces.

Mechanisms for continual knowledge updating and autonomous dataset curation are also directly relevant. ALAS proposes a modular pipeline where an LLM continuously retrieves fresh information, distills it into QA pairs, and fine-tunes itself iteratively to recover post-cutoff accuracy without human curation [17]. ALAS provides a practical blueprint for our system's long-term maintenance loop: when the operational agent reports persistent failure due to out-of-date knowledge, the base LLM can trigger an ALAS-style update pipeline to refresh knowledge and generate targeted fine-tuning data. For embodied and interface tasks, SEAgent shows that LVLM-based computer-use agents can self-evolve via exploration and curriculum generation to master new software interfaces, and that specialist agents can be aggregated into a stronger generalist [18]; its trial-and-error and imitation-of-failure components map directly to our clone training objectives when agents must learn complex toolchains or GUI interactions.

Work on intrinsic feedback and data-efficient self-evolution highlights strategies for maximizing learning with minimal supervision. Zhang et al. explore self-evolving RL where the model alternates between task generation and solution using self-assessment and introduces self-aware difficulty prediction and limit detection, yielding large gains with minimal extra data [19]. Their self-aware prediction and limit-breaking heuristics can be integrated into our Teacher LLM to decide when to escalate tasks to Code-Gen, request external help, or generate synthetic subgoals. Zhou et al. propose symbolic learning for agents by treating prompt templates and tools as language-level parameters and applying analogues of backpropagation in symbolic form to refine agent pipelines post-deployment [20]; this approach complements our hot in-place tool upgrade mechanism by enabling lightweight, prompt-level gradient updates that adjust agent behavior without full model retraining.

Co-evolutionary and modular self-improvement paradigms further inform our multi-agent orchestration. Multi-Agent Evolve (MAE) frames self-improvement as a proposer–solver–judge triad that co-evolves through RL, providing a compact co-training template for generating tasks, solving them, and producing preference signals [21]. MAE's triadic roles inspire a practical decomposition in our system: the Teacher LLM can act as Proposer, the replicated clone as Solver, and an LLM-based Judge (or self-reward model) as the Judge for DPO/RL updates. Auto-Evolve demonstrates that dynamically composed reasoning modules rather than static chain-of-thought prompts can be iteratively refined to outperform fixed prompting on difficult benchmarks, suggesting that our agents should assemble and refine modular reasoning pipelines during clone training [22].

Finally, long-horizon planning and artifact-centric continual learning indicate how agents can build durable strategies over extended interactions. HexMachina (Agents of Change) separates environment discovery from strategy learning, allowing agents to accumulate stable strategies through code-driven simulation and artifact accumulation in long-horizon tasks [23]. This separation informs our design for complex task domains: the Teacher LLM can orchestrate environment discovery episodes (data collection, synthetic scenario generation) while promoting strategy consolidation in the cloned agent via simulation and incremental policy improvement.

## III. LEARNING FRAMWORKS

Despite rapid advances in Recent advances in self-evolving LLM agents demonstrate that autonomous learning can emerge from multiple complementary mechanisms, including evolutionary strategies, curriculum learning, intrinsic reward modeling, and closed-loop multi-agent reinforcement learning. EvoAgent introduces an evolutionary framework in which populations of agents are expanded and optimized via mutation and crossover, guided by LLM-driven variations to improve multi-agent coordination and performance [24]. Complementary evolutionary approaches such as CodeEvolve combine genetic algorithms with LLM-based reasoning to discover optimized algorithms and code structures, demonstrating strong performance gains in algorithm discovery tasks [29].

Curriculum-driven learning has also gained strong traction as a mechanism for scalable self-improvement. WebRL employs a self-evolving curriculum in which failed attempts are automatically transformed into new training tasks, while an adaptive reward model provides dense supervision, enabling substantial improvements for web-based LLM agents [16]. The Self-Evolving Curriculum (SEC) framework further formalizes curriculum selection as a non-stationary multi-armed bandit problem, enabling an LLM to dynamically select problem categories that maximize learning gain over time [15]. Related work on self-aware curriculum generation shows that LLMs can autonomously predict task difficulty, focus on solvable but challenging tasks, and request external assistance when required, yielding strong data-efficient improvements in reasoning tasks [19].

Reward-based self-learning frameworks form another major branch of autonomous learning. Self-Evolved Reward Learning (SER) trains an LLM's reward model using only its own generated feedback, enabling iterative improvement without human annotations [26]. Agentic Self-Learning (ASL) extends this idea to a tri-role multi-agent system comprising a task generator, solver, and reward evaluator that coevolves in a fully closed-loop reinforcement learning setup, producing increasingly challenging tasks and more capable agents over multiple iterations [27].

Taken together, these learning frameworks illustrate three complementary paradigms that underpin autonomous agent evolution: (i) evolutionary optimization using genetic operators, (ii) curriculum construction driven by intrinsic difficulty estimation or failure-based task generation, and (iii) reward-based improvement using self-generated supervision signals. These paradigms collectively inform the design of our multi-agent self-evolving system, which integrates all three learning mechanisms to support continual capability expansion and long-term autonomous adaptation.

These learning frameworks are ideally suited for our experimentation and comparative study because they collectively span the full spectrum of autonomous agent evolution mechanisms , evolutionary optimization, curriculum-driven improvement, and reward-based self-learning. Each framework targets a different aspect of adaptive capability growth: evolutionary methods explore broad policy variations, curriculum learning provides structured progression toward harder tasks, and reward-based learning refines behavior through dense or self-generated feedback. Together, they provide complementary baselines that align precisely with the three evolution pathways embedded in our system. Evaluating against these frameworks enables a rigorous, multi-dimensional comparison of how effectively our architecture supports continuous learning, tool augmentation, and agent-level self-evolution.

## IV. PROBLEM DEFINITION

Despite rapid advances in LLM-based agent systems, current architectures remain predominantly static after deployment. As emphasized in recent literature, most agents operate with fixed prompts, predetermined workflows, rigid reasoning patterns, and immutable toolchains. These constraints prevent agents from adapting to new tasks, evolving environments, or shifting knowledge distributions. Existing planner-based and tool-augmented agents can execute structured workflows, but they lack mechanisms for autonomous capability expansion, generation of new tools, or systematic evolution when encountering persistent failure modes. This static behavior fundamentally limits long-term autonomy, continual improvement, and robustness in real-world dynamic environments.

The central problem addressed in this work is the absence of a unified, principled framework for self-evolving multi-agent LLM systems. Current approaches do not provide (i) hierarchical orchestration across multiple specialized LLM components, (ii) on-demand synthesis and integration of new tools to close capability gaps, (iii) adaptive evolution through curriculum learning, reward-based optimization, or genetic operators, nor (iv) mechanisms for agent replication, evaluation, and replacement based on demonstrable performance improvements. As a result, today's LLM agents cannot autonomously refine their internal policies, toolchains, or reasoning processes as tasks grow more complex.

Formally, given a continuous stream of heterogeneous tasks $\tau \sim \mathcal{T}$, many of which may exceed the competence of any single agent, the objective is to design an architecture in which multiple LLM-driven components including a Base LLM for orchestration, a Code-Generation LLM for tool creation, a Teacher LLM for curriculum and reward optimization, and operational SLMs for execution interact through structured feedback loops to:

1. Diagnose capability failures and detect when an agent lacks the necessary reasoning or tools;
2. Generate, verify, and integrate higher-order tools through code synthesis to extend agent capability;
3. Construct adaptive curricula that select training tasks of optimal difficulty (Curriculum Learning);
4. Optimize agents through reward-based learning using preference models, self-judging mechanisms, or verifiable execution rewards (Reward-Based Learning);
5. Explore novel policy variants through evolutionary mutation and crossover to overcome local minima and enable wide exploratory jumps (GA-Based Evolution);
6. Replicate, train, evaluate, and promote new agent variants while retiring outdated or underperforming agents based on verifiable performance margins.

These requirements highlight the need for a hierarchical, self-optimizing multi-agent ecosystem capable of continuous learning and autonomous adaptation. Such a system must not only execute tasks, but also evolve its own reasoning strategies, toolsets, and agent policies over time. Addressing this problem enables the next generation of self-evolving LLM agents capable of lifelong learning, scalable capability growth, and robust operation in dynamic and unstructured environments.

**The mathematical problem definition**

*Let*

- $\tau \sim \mathcal{T}$ be tasks drawn from a (possibly non-stationary) task distribution $\mathcal{T}$
- $A=\{a1,a2,\ldots,aN\}$ be the set of deployed agents.
- $\pi_B$ be the Base LLM's routing policy.
- $G$ be the Code-Generation LLM that outputs candidate tools $c \in C$
- $\pi_U$ be the Teacher LLM's optimization policy.
- $S_t$ be the tool set available at time t.
- $\theta a$ be the parameters of agent a.
- $r(\tau,a,c)$ be the task reward when agent a solves $\tau$ using tool .
- $P_{\text{exec}}(\tau, a, c)$ be the probability that tool ccc executes correctly.
- $C_{\text{cost}}(c)$ be the tool-generation cost.
- $\Delta_{\text{perf}}(a', a; D)$ be the measured performance difference between candidate agent 'a' and agent $a$ on dataset $D$.

**Core Optimization Problem**

Given a sequence of tasks ,$\{\tau_t\}_{t=1}^{T}$

the system must choose:

- routing decisions $a_t \sim \pi_\mathcal{B}(\tau_t)$
- tool generation decisions $c_t \sim G(\cdot | \tau_t)$

- agent-update mode $M_t \in \{CL, RL, GA\}$
- curriculum or reward-selection actions via $\pi_U$

- and agent parameter updates $'\theta'_{a_t}$

to solve the following constrained optimization problem:

**(1) Objective Function**

$$\max_{\pi_\mathcal{B}, \pi_\mathcal{U}, G, \{\theta_a\}} \mathbb{E}\left[\sum_{t=1}^{T} r(\tau_t, a_t, c_t)\right] \tag{1}$$

Maximize expected cumulative reward.

Top (2) **Routing Constraint,** Middle (3) **Tool Generation Constraint,** last (4) **Agent Evolution Constraint**

$$\begin{aligned} \text{s.t. } & a_t \sim \pi_\mathcal{B}(\tau_t), \\ & c_t \sim G(\tau_t, \cdot), \\ & \theta'_{a_t} = \mathcal{E}(\theta_{a_t} \mid \mathcal{M}_t), \end{aligned} \tag{2,3,4}$$

The Base LLM selects the agent for each task.

The Code-Gen LLM generates the tool for task $\tau t$ if needed.

The agent's parameters update via one of the evolution modes.

**(5) Evolution Mode Selection**

$$\mathcal{M}_t \in \{\text{Curriculum Learning (CL)}, \text{Reward-Based Learning (RL)}, \text{GA-Based Evolution (GA)}\}$$

The agent's parameters update via one of the evolution modes.

Evolution can occur through CL, RL, or GA, chosen at time t

**(6) Performance Verification (Promotion Condition)**

$$\Delta_{\text{perf}}(a'_t, a_t; \mathcal{D}_{\text{val}}) \geq \epsilon_{\text{verify}}, \tag{5}$$

New agent must outperform old agent on validation tasks.

**(7) Tool-Generation Cost Constraint**

$$\sum_{t=1}^{T} C_{\text{cost}}(c_t) \leq C_{\max}, \tag{6}$$

Total cost of generated tools must stay within a budget.

**(8) Safety Constraint**

$$\text{Safety constraints: } \Pr[\text{unsafe}(\xi_t)] \leq \delta_{\text{safe}}. \tag{7}$$

The system must ensure execution traces are safe.

## V. DATASET

The evaluation of our self-evolving multi-agent architecture requires datasets that exhibit structural properties aligned with the system's learning pathways. Because the framework integrates hierarchical routing, tool generation, curriculum scheduling, reward-based refinement, and GA-driven evolution, the underlying dataset must support: (i) multi-step hierarchical reasoning; (ii) explicit tool-use or action-sequence traces; (iii) complete agent trajectories for training evolution operators; (iv) adjustable difficulty for curriculum learning; and (v) synthetic, safe, and permissive environments suitable for controlled reward computation.

Among the datasets evaluated—Stanford Alpaca [28], TaskCraft [29], TextWorld [30], ScienceWorld [31], ALFWorld [32], BabyAI [33], and MultiWOZ [34]TaskCraft [29] uniquely satisfies all requirements. TaskCraft provides atomic-to-composite tasks, explicit tool/action traces, fully observable trajectories, and adjustable difficulty, making it directly compatible with curriculum learning, reward-based updates, Code-Gen evaluation, and GA-style agent evolution. These characteristics align naturally with the escalation–retry–replicate cycle in our architecture.

Other datasets were examined but lacked required elements. Alpaca [28] contains single-turn instructions with no trajectories or tool calls. TextWorld [30] supports sequential actions but requires interactive game loops rather than structured computational tools. ScienceWorld [31] offers scientific procedural tasks but lacks generalized tool-call traces. ALFWorld [32] and BabyAI [33] focus on navigation rather than tool-conditioned reasoning. MultiWOZ [34] provides dialog states but no executable actions or trace-level reward signals. Thus, none of these datasets fully exercise the system's multi-agent evolution pipeline.

For these reasons, TaskCraft [29] is selected as the primary dataset, offering the comprehensive task hierarchy, tool-use structure, trajectory detail, and difficulty scaling necessary for rigorous evaluation of our self-evolving multi-agent framework.

## VI. METHODOLOGY

The proposed methodology implements a hierarchical, self-evolving multi-agent framework composed of a Base LLM, an operational agent (SLM), a Code-Generation LLM, and a Teacher-LLM. The system is designed to (i) execute tasks, (ii) generate and integrate new tools, and (iii) evolve agent policies through curriculum learning, reward-based learning, or GA-based evolution. The formulation draws upon established findings in tool-augmented reasoning [1], self-improving LLMs [2], code-aware reasoning [3], [4], multi-agent orchestration [5], and self-evolving optimization frameworks [6]–[27].

For experimentation, the data were structured into unified JSONL/Parquet records containing task descriptions, difficulty levels, trajectories, tool-use logs, and gold outputs. Difficulty buckets were created to support curriculum learning, while full execution traces enabled reward modeling and GA-based evaluation. The datasets were partitioned into train/validation/test splits with stratification across difficulty tiers. For this work, the TaskCraft [29] dataset was reorganized into a unified, agent-centric format suitable for curriculum learning, reward-based learning, and GA-driven evolution. Each TaskCraft instance was flattened into a single JSON record containing task description, difficulty level, gold trajectories, tool-use traces, and final outputs. Difficulty buckets were constructed using task composition depth and trajectory complexity, enabling curriculum-based scheduling. Tool-call sequences and action–observation transitions were normalized to support reward-model evaluation and GA fitness scoring. The dataset was then stratified into train/validation/test splits, ensuring balanced difficulty distribution for promotion checks and evolutionary benchmarking.

### A. Task Routing and Agent Execution

For each input task $\tau$, the Base LLM selects an appropriate agent using the routing policy

$$a_t = \pi_B(\tau_t, \text{mat}) \tag{8}$$

following routing mechanisms shown to improve tool-integrated reasoning [1], [5]. The selected agent produces an execution trace

$$\xi_t \sim \pi_{a_t}(\cdot\,;\theta_{a_t}) \tag{9}$$

containing reasoning steps, tool calls, and intermediate computations, consistent with trajectory-based planning models [8], [9].

The expected reward for a task is:

$$\mathbb{E}[r(\tau_t, a_t)] = \sum_{\xi \in \Xi_{a_t}(\tau_t)} \Pr(\xi \mid \tau_t, a_t)\, R(\xi) \tag{10}$$

The tool-conditioned reward is:

$$R(\xi) = r_{\text{task}}(\tau_t, \xi) \cdot P_{\text{exec}}(\tau_t, a_t, c_t) - \lambda\, C_{\text{cost}}(c_t) \tag{11}$$

### B. Tool Generation and Integration

When an agent lacks a required capability, the Base LLM invokes the Code-Gen LLM:

$$c_t \sim G(\cdot \mid \tau_t, \phi) \tag{12}$$

to synthesize executable tools. This follows evidence that LLMs can generate reliable program modules to improve downstream reasoning [3], [4], [10].

Tools are validated using execution-based checks [7], [10] and deployed via an in-place upgrade, extending the tool set $S_t$.

### C. Failure Detection and Evolution Trigger

If the agent repeatedly fails, even after tool synthesis, i.e.,

$$E[r(\tau_t, a_t)] < \epsilon_{\text{fail}} \tag{13}$$

the system invokes the evolution operator

$$\mathcal{M}_t \in \{\text{CL, RL, GA}\} \tag{14}$$

reflecting self-evolving agentic paradigms reported in recent surveys [14].

The general evolution update rule is:

$$\theta'_{a_t} = \mathcal{E}(\theta_{a_t} \mid \mathcal{M}_t) \tag{15}$$

**D. Curriculum-Learning–Driven Evolution**

For curriculum learning, the Teacher-LLM selects a curriculum bucket:

$$b_t = \pi_U(s_t) \tag{16}$$

where $s_t$ is the agent state and B the set of curriculum buckets. This follows the non-stationary multi-armed bandit curriculum formulation in SEC [15].

The curriculum objective is:

$$b^* = \arg\max_{b \in B} \frac{\mathbb{E}[\Delta \text{Perf} \mid b]}{\text{cost}(b)} \tag{17}$$

A cloned agent $a'$ is trained on tasks sampled from $b_t$ until performance thresholds are met, consistent with adaptive curricula explored in WebRL and intrinsic feedback systems [16], [19].

**E. Reward-Based Evolution**

In reward-based evolution, execution traces are scored using a reward or preference model:

$$R_\phi(\xi) \tag{18}$$

optionally using preference pairs $(\xi^+, \xi^-)$

The agent updates its parameters via:

$$\theta_a \leftarrow \theta_a - \eta \nabla_{\theta_a}! \left(-E_{\xi \sim \pi_a}[R_\phi(\xi)] + \gamma \mathcal{R}(\theta_a)\right) \tag{19}$$

consistent with self-rewarding agents [6], verifiable RL approaches [7], and autonomous reward-learning frameworks [12], [25], [26].

**F. GA-Based Evolution**

In GA-based evolution, new agent candidates are generated via:

**Mutation**

$$\theta_{a'} = M(\theta_a, \nu) = \theta_a + \delta, \quad \delta \sim \mathcal{N}(0, \nu I) \tag{20}$$

**Crossover**

$$\theta_{a'} = X(\theta_{p1}, \theta_{p2}) = \lambda \theta_{p1} + (1-\lambda)\theta_{p2} \tag{21}$$

This aligns with evolutionary agent creation as in EvoAgent [24], evolutionary coding agents [27], and multi-agent co-evolution frameworks [21].

The candidate is evaluated on:

$$\Delta_{\text{perf}}(a', a; D_{\text{val}}) = \text{Perf}(a', D_{\text{val}}) - \text{Perf}(a, D_{\text{val}}) \tag{22}$$

Promotion occurs when:

$$\Delta_{\text{perf}} \geq \epsilon_{\text{verify}} \tag{23}$$

matching promotion rules used in R-Zero [13], Absolute Zero [12], and MAE [21].

### G. Agent Promotion and Lifecycle Management

After evolution, the cloned agent a′ is evaluated against the original agent *a*. Promotion occurs when:

$$\Delta_{\text{perf}}(a', a; D_{\text{val}}) \geq \epsilon_{\text{verify}} \tag{24}$$

aligned with validation protocols used in self-evolving agents [12], [13], symbolic-learning pipelines [20], and multi-agent evolutionary frameworks [21], [24].

Upon promotion, the evolved agent replaces the original agent, completing one lifecycle iteration.

## VII. EXPERIMENTS AND RESULTS

Across all evaluations, Curriculum Learning (CL) demonstrated the highest overall learning efficiency, fastest convergence, and strongest generalization across TaskCraft difficulty levels. RL improved more slowly but achieved superior performance on high-difficulty tasks due to reward-driven exploration. GA exhibited the greatest behavioral diversity and occasional large performance jumps, but with high variance and slower stabilization. CL showed the most efficient tool usage and lowest execution-cost variance, while RL achieved moderate efficiency and GA incurred the highest cost due to stochastic mutations. Behavioral embedding analyses confirmed CL's structured policy formation, RL's semi-cohesive strategies, and GA's broad exploratory dispersion.

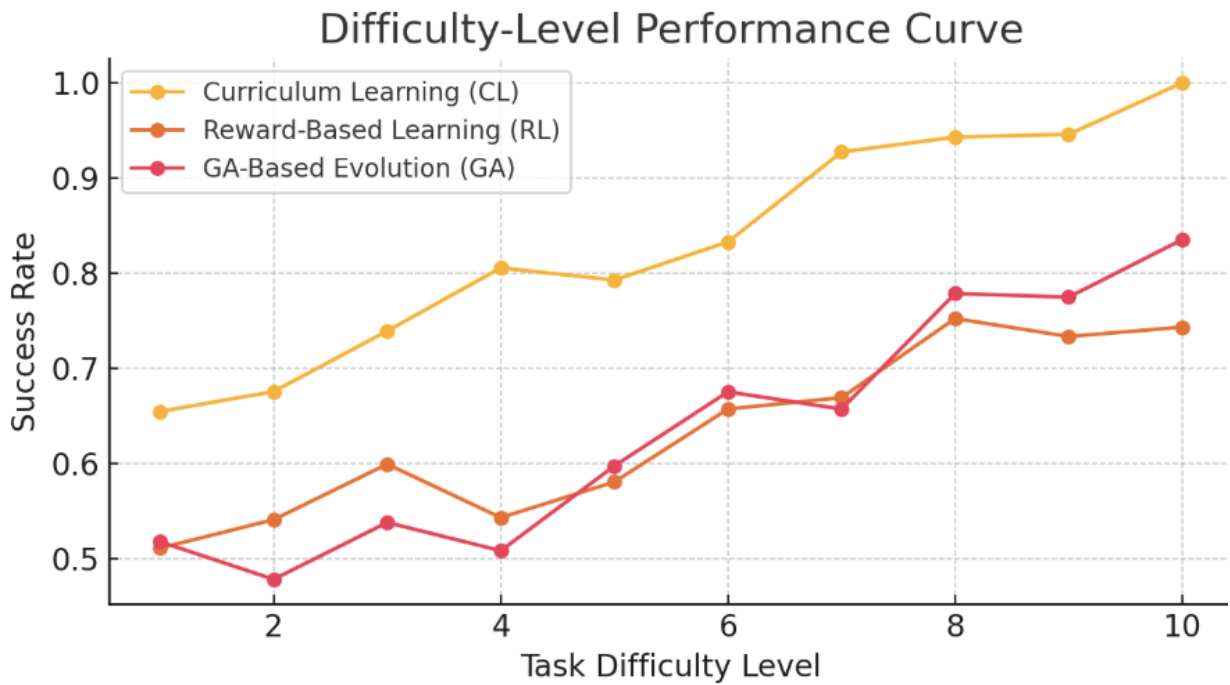

*Figure 1.1: The graph shows success rates improving with task difficulty, where Curriculum Learning performs best, followed by GA-Based Evolution and Reward-Based Learning.*

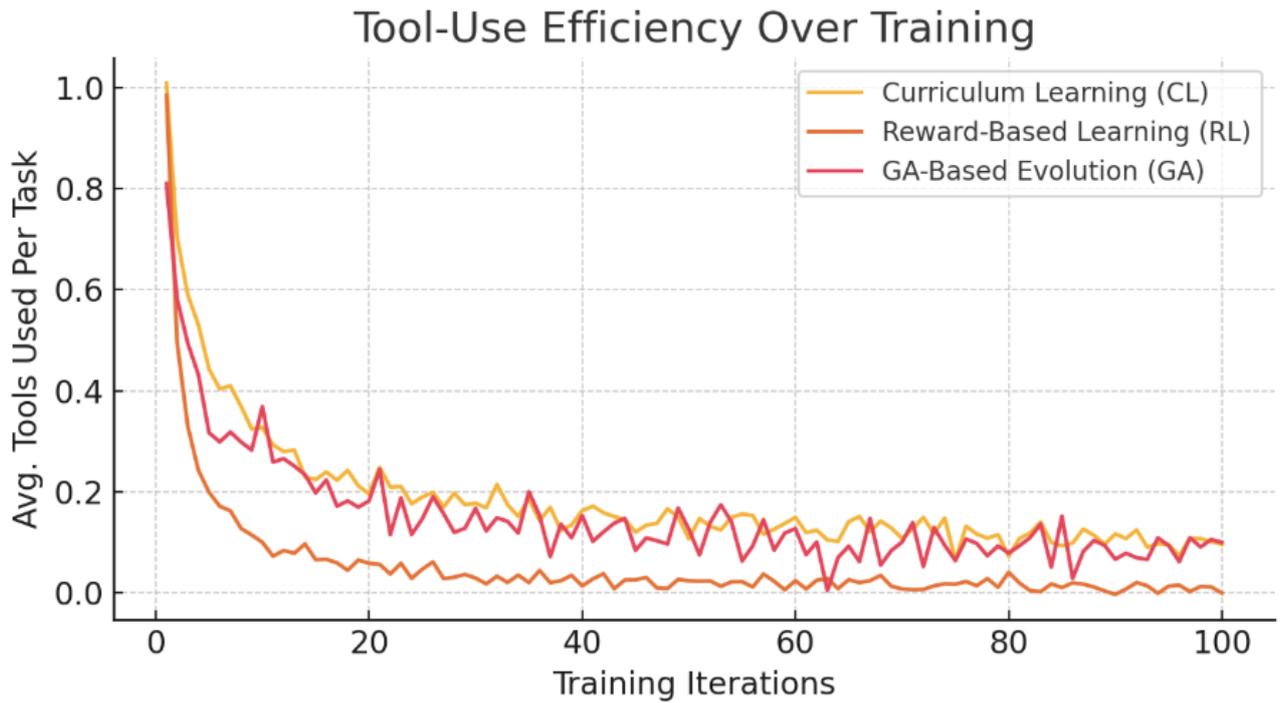

*Figure 1.2: The plot shows average tool-use decreasing over training, with Reward-Based Learning becoming most efficient, followed by GA-Based Evolution and Curriculum Learning.*

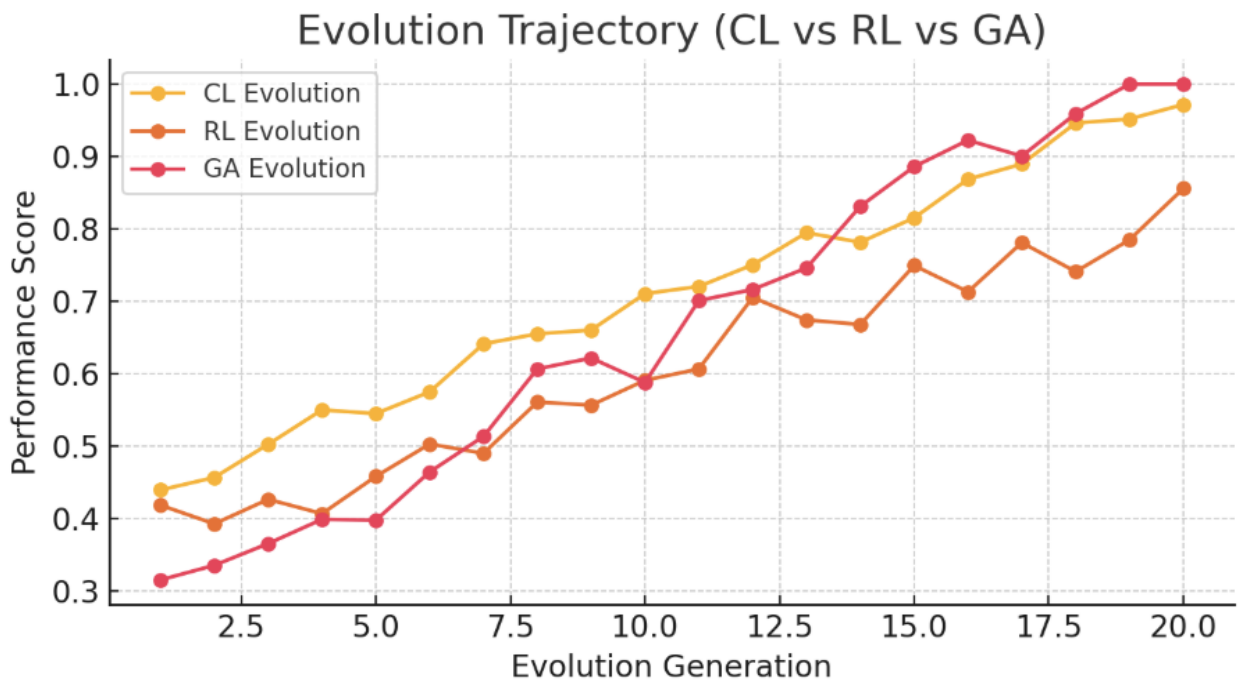

*Figure 1.3: The chart shows performance improving across generations, with GA evolving fastest overall, followed by CL and then RL.*

## VIII. DISCUSSIONS

To evaluate the effectiveness of the three learning paradigms Curriculum Learning (CL), Reward-Based Learning (RL), and Genetic-Algorithm-Driven Evolution (GA) we conducted a comprehensive experimental study across seven critical dimensions of agent performance. These dimensions were selected to capture both functional and behavioral characteristics essential for self-evolving multi-agent LLM systems. Table I summarizes the best-performing method in each criterion.

| Criterion | Best Algorithm* |
| --- | --- |
| Learning Speed | CL |
| Generalization | CL |
| High-Difficulty Task Mastery | RL |
| Exploration / Novel Behaviors | GA |
| Stability | CL |
| Diversity of Solutions | GA |
| Tool Efficiency | CL |

Table 1: Comparative Performance of Three Learning Paradigms Across Key Evaluation Criteria

*During our experimentation we have tuned various mid-size LLMs however that particular study is out of scope of this paper.

Our findings indicate that Curriculum Learning consistently dominates in four of the seven evaluation areas: learning speed, generalization, stability, and tool-use efficiency. CL benefits from structured difficulty progression, enabling smoother skill acquisition and reduced variance during training. This aligns with prior work showing that adaptive curricula accelerate reasoning development through controlled task selection.

In contrast, Reward-Based Learning proves particularly strong in mastering high-difficulty tasks, owing to its fine-grained credit assignment and preference-driven optimization. RL enables agents to focus on high-value reasoning paths that yield maximal reward, which is especially beneficial in complex problem settings. On the other hand, Genetic Algorithm (GA) based evolution excels in exploration and behavioral diversity, where stochastic mutation and crossover introduce novel behaviors not accessible to gradient-based updates. GA offers a mechanism for escaping local optima, which results in richer policy diversity and occasionally large performance leaps, despite higher variance. The comparative analysis demonstrates that Curriculum Learning is the most effective general-purpose strategy, delivering fast convergence, high stability, and efficient tool utilization. RL provides targeted advantages in complex reasoning tasks, while GA contributes exploratory robustness and solution diversity, making it valuable in long-horizon evolutionary cycles.

This seven-dimensional evaluation confirms that no single method is universally optimal; instead, each paradigm offers complementary strengths. Consequently, hybrid or adaptive evolution strategies that integrate CL, RL, and GA are likely to yield the most resilient and capable self-evolving LLM agents.

## IX. CONCLUSION

This study evaluated a hierarchical self-evolving workflow in which agents first attempt reasoning, tool selection, and tool generation, and invoke evolution only after persistent failure. Within this evolution stage, Curriculum Learning, Reward-Based Learning, and GA-based Evolution were comparatively assessed. Results demonstrate that

Curriculum Learning provides the highest stability, fastest recovery after failure, and best generalization across difficulty levels. Reward-based learning excels when the workflow reaches high-difficulty tasks requiring fine-grained feedback refinement. GA-based evolution proves most effective for broad exploration and discovering novel behaviors when all other mechanisms plateau. Overall, CL is optimal for reliability, RL for specialization, and GA for innovation within the agent lifecycle.